%% file: acl_arxiv.tex
\documentclass[11pt]{article}

\usepackage[preprint]{acl}
\usepackage{times}
\usepackage{latexsym}
\usepackage[T1]{fontenc}
\usepackage[utf8]{inputenc}
\usepackage{microtype}
\usepackage{inconsolata}
\usepackage{graphicx}
\usepackage{booktabs}
\usepackage{multirow}
\usepackage{amsmath}
\usepackage{amssymb}
\usepackage{xcolor}
\usepackage{colortbl}
\usepackage{listings}
\usepackage[most]{tcolorbox}

\definecolor{lstbg}{gray}{0.97}
\definecolor{brightbg}{HTML}{FFF4E5}
\definecolor{brightrule}{HTML}{D9534F}
\definecolor{blindedbg}{HTML}{EAF2FB}
\definecolor{blindedrule}{HTML}{4A7AB8}

\newtcolorbox{brightcase}[1]{
  enhanced, breakable,
  colback=brightbg, colframe=brightrule,
  boxrule=0.4pt, arc=2pt,
  left=4pt, right=4pt, top=4pt, bottom=4pt,
  fonttitle=\bfseries\footnotesize\sffamily,
  coltitle=white, colbacktitle=brightrule,
  title={\textsc{bright} \quad #1}
}

\newtcolorbox{blindedcase}[1]{
  enhanced, breakable,
  colback=blindedbg, colframe=blindedrule,
  boxrule=0.4pt, arc=2pt,
  left=4pt, right=4pt, top=4pt, bottom=4pt,
  fonttitle=\bfseries\footnotesize\sffamily,
  coltitle=white, colbacktitle=blindedrule,
  title={\textsc{blinded} \quad #1}
}

\newcommand{\benchname}{\textsc{KTD-Fin}}

\title{From Knowing to Doing: A Memory-Controlled Benchmark for LLM Trading Agents on Stock Markets}

\author{
 \textbf{Taojie Zhu}\textsuperscript{1,2,$\dagger$}
 \thanks{Work done during internship at Stepfun.},
 \textbf{Wentao Zhao}\textsuperscript{1,2,$\dagger$},
 \textbf{Rui Sun}\textsuperscript{2,$\spadesuit$},
 \textbf{Beidi Luan}\textsuperscript{2},
 \textbf{Jiacheng Lu}\textsuperscript{2,4},
\\
 \textbf{Sinuo Wang}\textsuperscript{2,5},
 \textbf{Jing Li}\textsuperscript{2},
 \textbf{Daxin Jiang}\textsuperscript{2},
 \textbf{Yonghong He}\textsuperscript{1,$\ddagger$},
 \textbf{Zuo Bai}\textsuperscript{2,3,$\ddagger$}
\\
\\
 \textsuperscript{1}Tsinghua University, 
 \textsuperscript{2}Stepfun, \textsuperscript{3}FinStep
\\
 \textsuperscript{4}Shanghai Jiao Tong University, \textsuperscript{5}Adelaide University
\\
\\
 \textsuperscript{$\dagger$}Equal contribution. \quad
 \textsuperscript{$\ddagger$}Corresponding author. \quad
 \textsuperscript{$\spadesuit$}Project Leader.
\\
 Email: \texttt{heyh@sz.tsinghua.edu.cn}, \texttt{baizuo@stepfun.com}
}

\begin{document}
\maketitle

\begin{abstract}
Evaluating whether large language model (LLM) agents can profit in capital markets is increasingly framed as end-to-end trading: place an agent in a historical market, let it trade, and measure portfolio returns. This setup is vulnerable to two evaluation failures. First, long backtests often overlap with the knowledge cutoffs of frontier LLMs, allowing memorized tickers, dates, prices, and market narratives to substitute for investment reasoning. Second, raw returns are a noisy proxy for stock-selection ability, since positive performance may come from market beta, style exposure, or favorable regimes rather than genuine alpha.

We introduce \benchname{} (\textbf{K}nowing-\textbf{T}o-\textbf{D}oing \textbf{Fin}ancial Benchmark), an end-to-end stock-market trading benchmark that addresses both issues. \benchname{} uses a data-side masking protocol to anonymize key identifiers and calendar information consistently across prompts and tools, separating historical market memory from investment decision-making. It also incorporates a Barra-style performance attribution framework that decomposes portfolio returns into market, style, and stock-selection alpha components.

Across ten frontier LLM agents evaluated on the Chinese A-share CSI300 over a 2024--2026 trading window, masking substantially changes agent rationales, pushing them away from company-specific narratives and toward anonymized factor-based reasoning. Attribution analysis further shows that LLM agents' cumulative returns under leakage-controlled evaluation are largely explained by passive market and style exposure, with limited evidence of persistent stock-selection alpha. These findings suggest that financial LLM benchmarks should evaluate not only whether an agent makes money, but also whether the source of returns reflects transferable investment skill. We release \benchname{} as a reproducible template for leakage-controlled and attribution-aware evaluation of LLM trading agents.
\end{abstract}

\input{section/introduction}
\input{section/related}
\input{section/benchmark}
\input{section/setup}
\input{section/discussion}

\input{section/conclusion}
\input{section/limitations}

\bibliography{custom}

\appendix
\input{section/appendix}

\end{document}

%% file: section/introduction.tex
\section{Introduction}
\label{sec:intro}

\begin{figure*}[!t]
\centering
\includegraphics[width=0.95\linewidth]{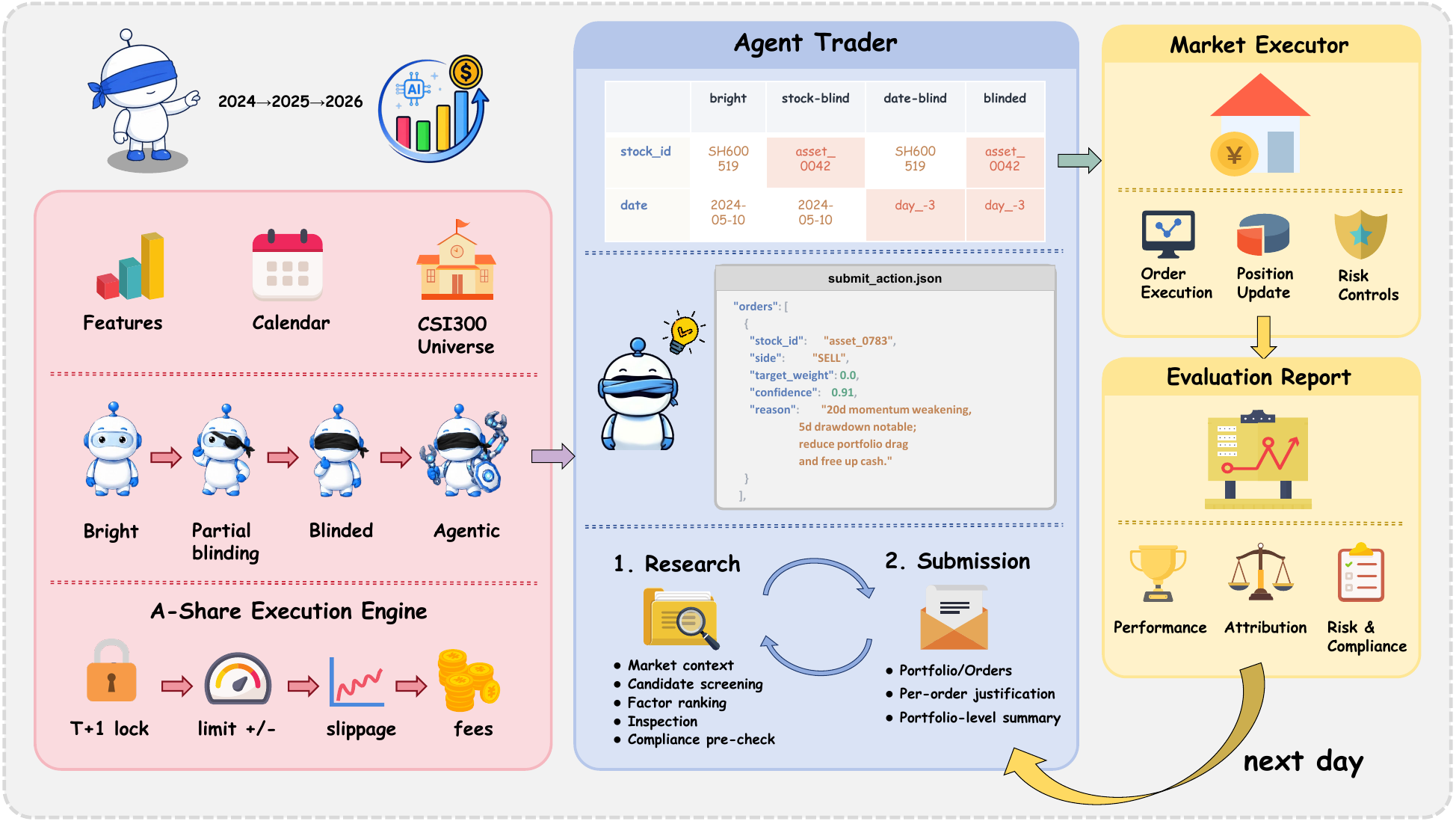}
\caption{Overview of \benchname{}. An LLM agent receives CSI300 features under one of four data-side masking conditions (\textit{bright}, \textit{stock-blind}, \textit{date-blind}, \textit{blinded}), executes a \textit{research}-then-\textit{submission} loop, and submits a JSON action that is run by a Qlib-based stock-market executor with realistic constraints.}
\label{fig:overview}
\end{figure*}

A growing number of recent systems treat the LLM not as a question-answerer about finance but as a trading agent: something that watches the market, decides what to look at next, and commits capital \citep{finmem2023,finagent2024,investorbench2024}. The benchmarks built around this shift fall into two groups. One keeps the model outside the market and tests how well it can read filings, answer questions, or extract structured information \citep{financebench2023,fineval2023,pixiu2023,finben2024}. The other puts the model inside a historical simulator, lets it trade, and reports portfolio return. As this end-to-end framing becomes the standard way to ask whether an LLM can ``profit'' in capital markets, two evaluation failures emerge. First, long backtests routinely overlap with the knowledge cutoffs of frontier models, so memorized tickers, prices, dates, and macro narratives can substitute for the data the benchmark thinks it is showing. Second, even when leakage is controlled, cumulative return is a noisy proxy for stock-selection skill: positive performance may simply reflect market beta, persistent style exposure, or a favorable regime rather than anything the agent has decided.

Take the first failure concretely. Real tickers and dates are passed to the model verbatim, and pretraining corpora are full of price commentary, brand descriptions, and macro coverage \citep{deepfund2025,oracleproto2026}. In a sanity check on our own environment---where the agent sees only OHLCV-derived numbers---a representative model writes \emph{``\textsc{sh601857}---defensive blue-chip with strong downside protection, increase position''} when tickers are visible, but declines to trade in all 25 of 25 cells when those tickers are anonymized and the same numbers are fed. The signal driving the first behaviour is the ticker, not the data. Existing fixes are partial: DeepFund \citep{deepfund2025} restricts evaluation to post-cutoff dates, which shrinks the window with every model release and prevents shared windows across models; OracleProto \citep{oracleproto2026} shows in a static forecasting setting that prompt-level ``do not use external memory'' instructions fail, motivating data-level mitigation, which we extend from their static setting to sequential trading.

The second failure is subtler. A leakage-controlled backtest still collapses to a single number---cumulative return---that lumps market drift, persistent style tilts, and genuine stock-specific skill into one figure. Empirical asset pricing handles this by decomposing returns into market, style, and a stock-specific residual; without that decomposition, a financial LLM benchmark cannot distinguish ``the agent made money'' from ``the agent rode a factor any similarly-exposed portfolio would have ridden.''

\benchname{} (\textbf{K}nowing-\textbf{T}o-\textbf{D}oing \textbf{Fin}ancial Benchmark; Figure~\ref{fig:overview}) is a Qlib-based stock-market trading benchmark---instantiated on the CSI300 universe of the Chinese A-share market---designed to address both failures inside the protocol. For leakage, a four-level data-side \emph{masking protocol}---\textit{bright}, \textit{stock-blind}, \textit{date-blind}, \textit{blinded}---anonymizes tickers, calendar dates, and tool-return timestamps in coordination, so an agent in \textit{blinded} never sees a real identifier even transiently through a tool result; an independent ten-attacker probe (Section~\ref{sec:abl-deanon}) certifies the mask, with top-1 ticker recovery at most $3.0\%$ and a strict joint (ticker top-5 \emph{and} date within $\pm 7$ trading days) rate at most $1.5\%$. For the return-versus-skill conflation, \benchname{} reports a Barra-style daily cross-sectional attribution that decomposes each agent's realized return into market, style, and stock-selection alpha components. Three decision modes---\textit{memory-only}, \textit{fixed-candidate}, and \textit{open-research}---vary memory exposure and information access; reporting is consolidated into a ten-dimensional metric panel covering returns, behavior, and reliability.

The headline evaluation runs ten frontier LLMs in the cleanest \textit{blinded $\times$ open-research} setting over a single contiguous 2024-01-01 to 2026-04-10 window, against eighteen Alpha9-trained machine-learning baselines (trained on 2015--2022, validated on 2023) and CSI300 buy-and-hold. Two findings emerge that map directly onto the two failures we set out to address. (i) Masking changes the \emph{shape} of trading, not just the score: under \textit{bright}, agent rationales are brand-driven (``defensive blue-chip,'' ``liquor leader'') and overweight memorized index heavyweights; under \textit{blinded}, the same model switches to factor-rank reasoning over anonymized IDs. (ii) Cumulative return is not the same as stock-selection skill: the Barra attribution shows that LLM agents' headline returns are largely explained by passive exposure to market and style factors, with only one model at near-zero stock-selection alpha and the rest measurably negative.

\paragraph{Contributions.} (1) \benchname{}, an end-to-end stock-market trading benchmark that handles both failure modes inside the protocol: leakage via a four-level data-side masking protocol (certified by a ten-attacker de-anonymization probe) and return-versus-skill via a Barra-style cross-sectional attribution. (2) A headline ten-LLM evaluation under \textit{blinded $\times$ open-research} against eighteen Alpha9-trained baselines and CSI300 buy-and-hold, scored on a ten-dimensional metric panel and re-scored by Barra-attributed selection alpha. (3) Public release of environment, scripts, masking and probe protocols, and the full evaluation dataset.

%% file: section/related.tex
\section{Related Work}
\label{sec:related}

\paragraph{Benchmarking LLMs in Financial Domain.}
With the rapid growth of LLMs in finance, a substantial body of benchmarks has been developed to evaluate model capabilities in financial contexts \citep{financebench2023,fineval2023,pixiu2023,finben2024}. Financial LLM benchmarks have evolved from document-understanding frameworks such as FinanceBench \citep{financebench2023}, PIXIU/FLARE \citep{pixiu2023}, FinBen \citep{finben2024}, FinQA \citep{finqa2021}, ConvFinQA \citep{convfinqa2022}, TAT-QA \citep{tatqa2021}, and the Chinese-language CFLUE \citep{cflue2024} and FinEval \citep{fineval2023}, to trading-agent evaluations such as FinMem \citep{finmem2023}, FinAgent \citep{finagent2024}, InvestorBench \citep{investorbench2024}, and the multi-agent designs of TradingGPT \citep{tradinggpt2023}, FinCon \citep{fincon2024}, and CryptoTrade \citep{cryptotrade2024}, with open-source LLM and platform releases by FinGPT \citep{fingpt2023} and FinRobot \citep{finrobot2024}. These benchmarks share a focus on financial knowledge and trading decisions, but typically pass real tickers and trading calendars to the model, which conflates a model's reasoning with priors memorized during pretraining and leaves no controlled way to separate the two. Our work introduces a four-level data-side masking protocol that turns memory exposure into a controllable axis without depending on each model's training cutoff.

\paragraph{Contamination-aware Benchmarking.}
There has been growing interest in contamination-controlled evaluation. General-domain efforts include LiveCodeBench \citep{livecodebench2024} for rolling post-cutoff problem collection and Min-K\% Prob \citep{minkprob2024} for membership inference on pretraining data. In the financial domain, DeepFund \citep{deepfund2025} evaluates each model only on dates after its own training cutoff, and OracleProto \citep{oracleproto2026} controls exposure at the data layer for static forecasting and shows that prompt-level restrictions cannot substitute for it. These approaches either shrink the available evaluation window with every new model release or remain limited to single-step forecasting tasks. Our work is the first sequential-decision financial benchmark to control pretraining-memory exposure inside the protocol, with the masking independently certified by a ten-attacker de-anonymization probe.

\paragraph{Agent and infrastructure foundations.}
The agent loop in \benchname{} builds on ReAct \citep{react2023}, with related tool-use methods in Toolformer \citep{toolformer2023}, ToolLLM \citep{toolllm2024}, and HuggingGPT \citep{hugginggpt2023}, and general agent evaluation in AgentBench \citep{agentbench2024} and $\tau$-bench \citep{taubench2024}. The execution environment follows Qlib conventions \citep{qlib2020}; FinRL-Meta \citep{finrlmeta2022} and TradeMaster \citep{trademaster2023} provide adjacent RL-based trading platforms. Our gradient-boosted factor baseline uses LightGBM \citep{lightgbm2017}.

%% file: section/benchmark.tex
\section{Benchmark Design}
\label{sec:benchmark}

\benchname{} evaluates LLMs as stock-market trading agents under tight control of pretraining-memory exposure (Figure~\ref{fig:overview}). It has four interlocking components---a four-level masking protocol (§\ref{sec:masking}), three decision modes (§\ref{sec:modes}), a Qlib-based execution engine (§\ref{sec:market-env}), and a ten-dimensional metric panel (§\ref{sec:metrics})---each described below.

\subsection{Execution engine}
\label{sec:market-env}
The task is daily portfolio construction over the CSI300 constituent universe. At each trading step the agent observes account state, current positions, last-step execution results, benchmark-relative performance, and summary constraints, and must finally emit a JSON action. Observations are \emph{price-only}: OHLCV histories and derived technical features. The benchmark contains no news, analyst reports, regulatory filings, or fundamentals---any reasoning that names brand, sector, or fundamentals must therefore reflect pretraining priors rather than observation.

The execution engine simulates A-share trading at realistic granularity: long-only positions starting from CNY 1{,}000{,}000; T+1 sell availability; CNY 5 minimum cost per order; 5 bps buy cost and 15 bps sell cost. Orders execute at the next-day open price. Daily price-limit filtering excludes limit-up stocks from \texttt{BUY} candidates and limit-down stocks from \texttt{SELL} candidates, using sector-differentiated thresholds that match the A-share board rules (main board $\pm 10\%$, ChiNext/STAR $\pm 20\%$, BSE $\pm 30\%$; full board-code mapping in Appendix~\ref{sec:appendix-qlib}). The simulator logs all violations and execution outcomes so invalid LLM actions remain visible rather than being silently corrected.

\subsection{Three decision modes}
\label{sec:modes}
\benchname{} decomposes LLM trading competence along three orthogonal capability dimensions, each instantiated as an independent decision mode (Table~\ref{tab:modes}).

\begin{table}[!htbp]
\centering
\footnotesize
\setlength{\tabcolsep}{4pt}
\begin{tabular}{@{}p{1.9cm}p{2.3cm}p{2.7cm}@{}}
\toprule
Mode & Capability probed & Non-prior information \\
\midrule
memory-only     & prior-memory dependence    & stock IDs only \\
fixed-candidate & clean-signal selection     & curated factor table \\
open-research       & end-to-end agentic reasoning & six retrieval tools \\
\bottomrule
\end{tabular}
\caption{Three decision modes, one per capability dimension. All modes share identical execution constraints, masking, and account state.}
\label{tab:modes}
\end{table}

The memory-only mode carries only stock identifiers and account state, so any non-trivial trade must come from pretraining priors over the listed IDs; fixed-candidate hands the agent a curated OHLCV+factor table over a small factor-top-$k$ union (the typical prior-work setting, probing structured-signal selection); open-research exposes a six-tool retrieval interface that the agent invokes freely (Section~\ref{sec:tools}), the fully autonomous agent setting.

\subsection{Four-level masking protocol}
\label{sec:masking}
The masking protocol is the data-side instrument by which \benchname{} controls pretraining-memory exposure. It operates at four levels: \textit{bright} (real ticker, real date), \textit{stock-blind} (anonymized ticker, real date), \textit{date-blind} (real ticker, relative day index), and \textit{blinded} (both anonymized). Figure~\ref{fig:overview} shows a concrete alias example. The two intermediate levels jointly form the partial blinding stratum that isolates ticker-specific versus calendar-specific priors. Aliases are stable within an episode and shuffled across episodes to prevent within-run identity reconstruction.

The benchmark harness preserves end-to-end masking consistency. An episode-level alias map translates real tickers and dates to aliases when constructing the prompt; tool arguments are un-masked before querying the data layer and re-masked before being returned to the model. Therefore an agent in blinded mode never observes a real ticker or a real calendar date, even transiently through a tool result. All numeric features the model receives---returns, volatilities, drawdowns---are computed on the original time series so the underlying signal is preserved; only the identifying handles change.

\paragraph{Mask validity.} An independent ten-attacker de-anonymization probe (Section~\ref{sec:abl-deanon}) finds joint ticker+date recovery at most $1.5\%$, confirming that residual leakage through board, sector, or volatility-profile features is negligible; in what follows we treat blinded as a near-clean reasoning condition.

\subsection{Tool-mediated open research}
\label{sec:tools}
The open-research decision mode operates as a ReAct-style \citep{react2023} loop with a strict two-phase structure. In a research phase, the model issues calls---as many as it deems necessary---to six read-only tools covering market context, candidate screening, per-stock snapshots, pairwise comparison, portfolio state, and pre-trade risk checks (full interface in Appendix~\ref{sec:appendix-tools}, Table~\ref{tab:tools}). The initial prompt contains only account state and constraints, so the model must actively retrieve any market information through tools. In a submission phase, the model is forced via \texttt{tool\_choice} to call \texttt{submit\_action} and emit a JSON order list, with up to three retries on schema or constraint violations and a fallback to no-trade if all retries fail. Failure modes during the research phase (parse errors) inject structured feedback rather than aborting the run. The harness is provider-neutral: provider-specific tool-call wire formats (OpenAI-compatible, Anthropic-native) are normalized inside per-provider adapters.

\subsection{Action schema}
\label{sec:action-schema}
The final action is a JSON object with an \texttt{orders} list and an \texttt{overall\_reason} string. Each order contains \texttt{stock\_id}, \texttt{side} (\texttt{BUY} or \texttt{SELL}), \texttt{confidence} in $[0,1]$, a short \texttt{reason}, and exactly one of \texttt{target\_weight} or \texttt{shares}. The executor rejects orders that violate universe membership, position limits, T+1, or formatting constraints; rejected orders are logged but do not abort the step. The \texttt{confidence} field feeds the calibration metrics; the \texttt{reason} field is used in our motivation analysis (Section~\ref{sec:intro}).

\subsection{Ten-dimensional metric panel}
\label{sec:metrics}
A core thesis of \benchname{} is that LLM trading agents must be judged not only on what they earn but on how they earn it. We report a ten-dimensional metric panel grouped into three families: returns and risk (total return, Sharpe, max drawdown, information ratio); behavior (annualized turnover, Herfindahl--Hirschman index, cash ratio, abstention rate); and reliability and calibration (parse-failure rate, expected calibration error). Two further quantities---tool-validity rate and the Brier score on per-order \texttt{confidence}---are recorded for diagnostic use but not part of the headline panel. Full definitions are in Appendix~\ref{sec:appendix-metrics}.

%% file: section/setup.tex
\section{Experiments}
\label{sec:experiments}
\label{sec:setup}

\begin{figure*}[!t]
\centering
\includegraphics[width=\linewidth]{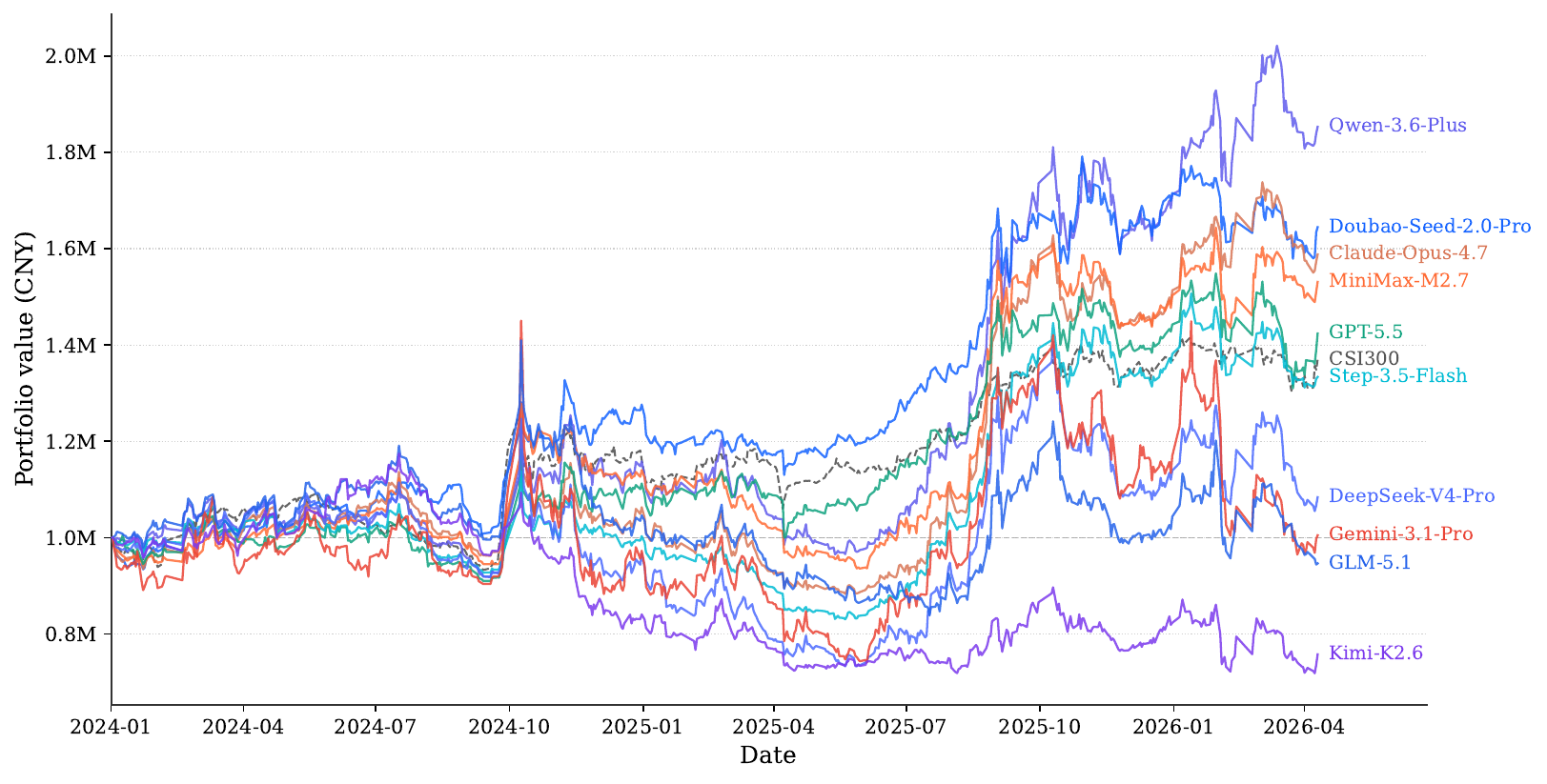}
\caption{Seed-averaged daily portfolio equity curves of the ten LLM agents and CSI300 buy-and-hold (dashed grey) under \textit{blinded $\times$ open-research} on the long 2024-01-01 to 2026-04-10 window, starting from CNY $1{,}000{,}000$.}
\label{fig:equity-curves}
\end{figure*}

\subsection{Setup}
\label{sec:setup-section}

\paragraph{Evaluation windows.}
We evaluate on ten regime-stratified short windows of $\sim$43 trading days each (W1--W10, Table~\ref{tab:windows}) plus one contiguous long window from 2024-01-01 to 2026-04-10 (548 trading days). Only W10 falls strictly post-cutoff for the anchor model; the other nine rely on the masking protocol (Section~\ref{sec:masking}) for contamination control. The long window drives the headline cross-model comparison; the short windows drive per-regime analysis on the anchor.

\begin{table}[!htbp]
\centering
\footnotesize
\setlength{\tabcolsep}{4pt}
\begin{tabular}{@{}lp{2.8cm}rl@{}}
\toprule
Code & Dates & CSI300 & Regime \\
\midrule
W1   & 2024-04-01--06-05      & $-0.02\%$ & near-flat \\
W2   & 2024-05-27--07-25      & $-6.50\%$ & monotone bear \\
W3   & 2024-07-25--09-25      & $+0.07\%$ & near-flat \\
W4   & 2024-09-09--11-15      & $+24.30\%$ & explosive bull \\
W5   & 2024-10-21--12-13      & $-0.05\%$ & near-flat \\
W6   & 2024-12-06--2025-02-14 & $-0.86\%$ & mild decline \\
W7   & 2025-03-03--05-09      & $-1.09\%$ & V-shape \\
W8   & 2025-07-14--09-19      & $+12.05\%$ & smooth uptrend \\
W9   & 2025-10-13--12-19      & $-0.56\%$ & near-flat \\
W10$^{\dagger}$ & 2026-02-02--04-10 & $+0.66\%$ & mild rise \\
\bottomrule
\end{tabular}
\caption{Ten regime-stratified evaluation windows. $^{\dagger}$W10 is strictly post-cutoff for the anchor model.}
\label{tab:windows}
\end{table}

\paragraph{Models.}
\label{sec:setup-models}
The panel separates depth from breadth. The anchor Step-3.5-Flash sweeps the full mask $\times$ mode grid (4 masks $\times$ 3 decision modes $\times$ W1--W10 $\times$ 5 seeds, with informative-cell pruning) to probe mask, mode, and seed effects. Ten frontier LLMs then run the headline comparison under \textit{blinded $\times$ open-research} on the long window---one model each from Step, OpenAI, Anthropic, Google, ByteDance, DeepSeek, Alibaba, MiniMax, Zhipu, and Moonshot (full identifiers in Table~\ref{tab:ten-llm}). All ten share identical prompts, tool interfaces, masking, and execution rules (Section~\ref{sec:benchmark}), so cross-model gaps reflect model differences alone. Each is run with multiple seeds; cells report the median.

\paragraph{Baselines.}
We benchmark the ten LLM agents against eighteen trained Qlib factor models on the long window. The baselines share the nine VIF-screened Alpha9 features (Table~\ref{tab:barra-factors}), are trained on 2008--2022 with 2023 as validation, and execute through the same trading layer as the LLM agents. They span linear, gradient-boosted trees, tabular networks, RNN/Transformer time-series models, and specialized time-series architectures (full model list and Qlib default hyperparameters in Appendix~\ref{sec:appendix-qlib}). CSI300 buy-and-hold is reported as the market reference.

\paragraph{Protocol.}
Multi-seed cells report median total return with $[Q_1, Q_3]$ as IQR; cross-condition contrasts use Wilcoxon signed-rank tests on paired (window, seed) medians at $\alpha = 0.05$. Mask validity is verified independently (Section~\ref{sec:abl-deanon}). The harness is deterministic given a seed, so any model is re-evaluable on identical conditions. LLM agents run via vendor APIs; the eighteen Qlib models train on one H800 (5\,min to $\sim 4$\,h). One LLM on the long window with three seeds takes 6--12\,h wall-clock.

\subsection{Anchor depth scan}
\label{sec:setup-stages}

On the anchor Step-3.5-Flash we sweep the four masks under each of the three decision modes (Table~\ref{tab:modes}) and additionally collect order-level rationales under \textit{bright} to characterize how the model talks when real identifiers are visible. Table~\ref{tab:mode-bright-blinded} summarizes the resulting mode $\times$ mask profile on the anchor depth set. Two patterns emerge: pretraining memory of real tickers alone is sufficient to drive active trading even when no observation channel is present---under \textit{memory-only}, \textit{bright} produces non-trivial orders while \textit{blinded} produces voluntary cash holding at exactly $0.00\%$; and granting the agent tool access (\textit{open-research}) consistently lifts capability above the no-tool \textit{fixed-candidate} baseline at both mask levels.

\begin{table}[!htbp]
\centering
\footnotesize
\setlength{\tabcolsep}{6pt}
\begin{tabular}{@{}lrr@{}}
\toprule
Mode & bright & blinded \\
\midrule
open-research   & $\mathbf{-2.11}$ & $\mathbf{-1.83}$ \\
fixed-candidate & $-3.61$          & $-3.76$ \\
memory-only     & $-0.16$          & $\phantom{-}0.00$ \\
\bottomrule
\end{tabular}
\caption{Median total return (\%) on the anchor depth set $\times$ 5 seeds by mode and mask. Bold: best per column.}
\label{tab:mode-bright-blinded}
\end{table}

\subsection{De-anonymization probe}
\label{sec:abl-deanon}

\begin{figure}[!htbp]
\centering
\includegraphics[width=0.95\linewidth]{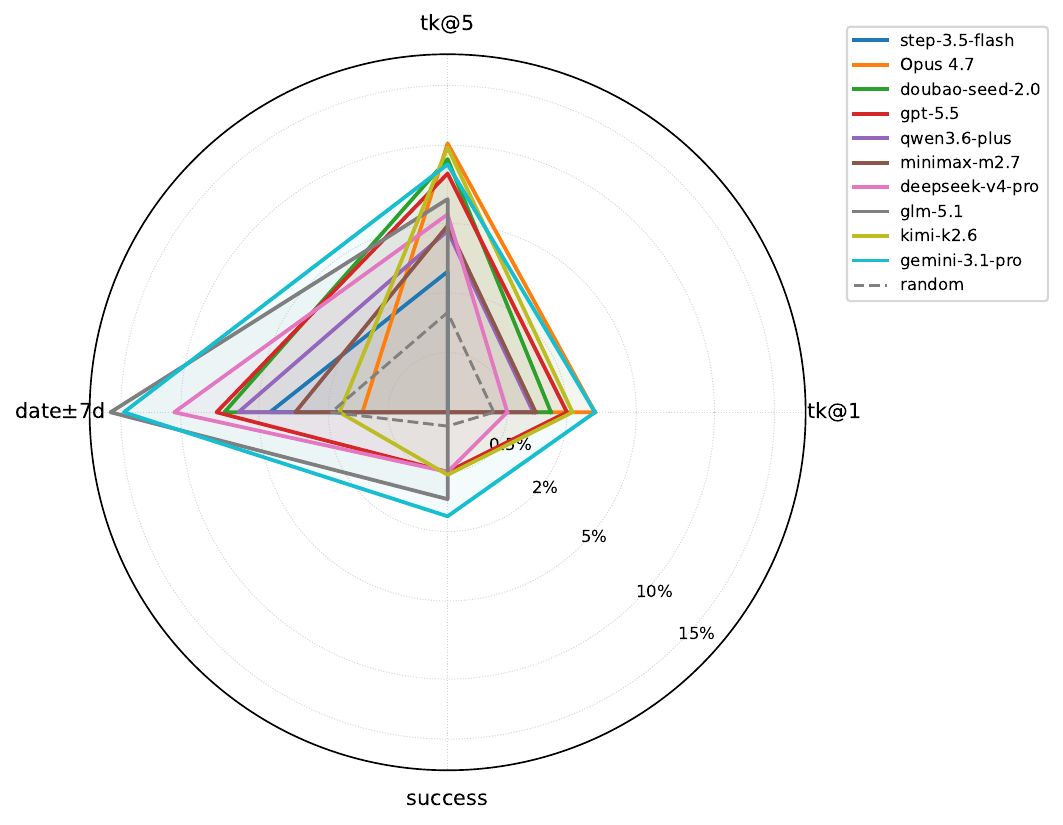}
\caption{Per-attacker recovery under \textit{blinded} on tk@1, tk@5, date $\pm 7$d, and joint success (sqrt-scaled). Grey dashed: random baseline.}
\label{fig:deanon-radar}
\end{figure}

To certify the masking protocol independently of the trading benchmark, we run a de-anonymization probe in which the exact masked payload that the agent sees in \textit{blinded} mode is fed to ten frontier LLMs acting as attackers. Each attacker is asked to recover the ticker (top-5 list), date, sector, and board for 200 (stock, date) probes stratified across five time windows and four boards; we score top-1 / top-5 ticker accuracy, date-within-$\pm 7$ trading-day rate, and a strict joint success rate (ticker top-5 \emph{and} date within $\pm 7$ trading days). Figure~\ref{fig:deanon-radar} shows the per-attacker recovery profile; the strongest attacker recovers only $10.2\%$ ticker top-5 and joint success peaks at $1.5\%$, both within a small additive margin of the random baseline---confirming that \textit{blinded} effectively cuts the model off from its pretraining memory of the underlying identifiers.

\subsection{Cross-model leaderboard}
\label{sec:results-headline}
\label{sec:results-open-research}

Under \textit{blinded $\times$ open-research} on the long 2024-01-01 to 2026-04-10 window, Figure~\ref{fig:equity-curves} shows the seed-averaged daily equity curves of the ten LLM agents against CSI300; Table~\ref{tab:ten-llm} reports the ten-dimensional metric panel for the LLMs alongside the eighteen supervised Qlib factor models, with CSI300 buy-and-hold inserted into each panel at its return-implied position.

\begin{table*}[!t]
\centering
\footnotesize
\setlength{\tabcolsep}{4pt}
\resizebox{\textwidth}{!}{%
\begin{tabular}{lrrrrrrrrr}
\toprule
Model & Total ret. & Excess & Sharpe & MDD & IR & Turn. & HHI & ECE & $\alpha$-rk \\
\midrule
\multicolumn{10}{l}{\textit{LLM agents, blinded $\times$ open-research, long window 2024-01-01 to 2026-04-10}} \\
\texttt{qwen3.6-plus}       & $+85.29\%$ & $+48.37\%$ & $1.13$  & $-24.49\%$ & $+0.48$ & $31.88$  & $0.005$ & $0.367$ & $2$ \\
\texttt{gpt-5.5}            & $+61.26\%$ & $+24.34\%$ & $0.98$  & $-24.63\%$ & $+0.29$ & $14.38$  & $0.037$ & $0.272$ & $6$ \\
\texttt{doubao-seed-2.0}    & $+61.01\%$ & $+24.09\%$ & $1.15$  & $-15.62\%$ & $+0.29$ & $14.18$  & $0.004$ & $0.425$ & $3$ \\
\texttt{claude-opus-4-7}    & $+58.80\%$ & $+21.88\%$ & $0.98$  & $-26.69\%$ & $+0.26$ & $31.75$  & $0.012$ & $0.246$ & $\mathbf{1}$ \\
\texttt{minimax-m2.7}       & $+56.31\%$ & $+19.39\%$ & $0.89$  & $-29.43\%$ & $+0.25$ & $43.68$  & $0.029$ & $0.303$ & $4$ \\
\rowcolor{gray!18}
CSI300 buy-and-hold         & $+36.92\%$ & $0.00\%$   & ---     & ---        & ---     & $0.00$   & --- & --- & --- \\
\texttt{step-3.5-flash}     & $+21.86\%$ & $-15.06\%$ & $0.49$  & $-28.99\%$ & $-0.13$ & $24.66$  & $0.012$ & $0.357$ & $5$ \\
\texttt{gemini-3.1-pro}     & $+8.25\%$  & $-28.67\%$ & $0.29$  & $-49.67\%$ & $-0.11$ & $101.23$ & $0.066$ & $0.387$ & $7$ \\
\texttt{deepseek-v4-pro}    & $+1.39\%$  & $-35.53\%$ & $0.19$  & $-42.35\%$ & $-0.29$ & $73.49$  & $0.026$ & $0.352$ & $8$ \\
\texttt{glm-5.1}            & $-3.55\%$  & $-40.47\%$ & $0.10$  & $-44.88\%$ & $-0.40$ & $100.25$ & $0.032$ & $0.322$ & $9$ \\
\texttt{kimi-k2.6}          & $-24.23\%$ & $-61.15\%$ & $-0.34$ & $-45.49\%$ & $-0.77$ & $85.26$  & $0.090$ & $0.306$ & $10$ \\
\midrule
\multicolumn{10}{l}{\textit{Trained Qlib factor models (Alpha9, 2008--2022 train, 2023 val); 8 representatives, full 18 in Appendix~\ref{sec:appendix-qlib}}} \\
\texttt{SFM}            & $+86.58\%$ & $+49.66\%$ & $2.02$ & $-7.41\%$  & $+0.57$ & $22.47$  & --- & --- & --- \\
\texttt{DoubleEnsemble} & $+73.66\%$ & $+36.74\%$ & $1.41$ & $-18.24\%$ & $+0.41$ & $72.84$  & --- & --- & --- \\
\texttt{CatBoost}       & $+72.97\%$ & $+36.05\%$ & $1.43$ & $-16.74\%$ & $+0.40$ & $29.96$  & --- & --- & --- \\
\texttt{LightGBM}       & $+66.22\%$ & $+29.30\%$ & $1.29$ & $-15.59\%$ & $+0.33$ & $46.18$  & --- & --- & --- \\
\texttt{Transformer}    & $+60.35\%$ & $+23.43\%$ & $1.61$ & $-11.29\%$ & $+0.27$ & $11.22$  & --- & --- & --- \\
\texttt{LSTM}           & $+43.92\%$ & $+7.00\%$  & $1.12$ & $-11.76\%$ & $+0.06$ & $78.03$  & --- & --- & --- \\
\rowcolor{gray!18}
CSI300 buy-and-hold     & $+36.92\%$ & $0.00\%$   & ---    & ---        & ---     & $0.00$   & --- & --- & --- \\
\texttt{MLP}            & $+34.08\%$ & $-2.84\%$  & $0.69$ & $-21.75\%$ & $-0.01$ & $18.48$  & --- & --- & --- \\
\texttt{Linear}         & $+29.60\%$ & $-7.32\%$  & $0.74$ & $-13.88\%$ & $-0.12$ & $15.61$  & --- & --- & --- \\
\bottomrule
\end{tabular}%
}
\caption{Headline leaderboard, sorted by total return; the shaded CSI300 row sits at its rank position in each panel. $\alpha$-rk ranks LLMs by Barra-attributed selection alpha; bold marks the only agent with $\alpha \geq 0$.}

\label{tab:ten-llm}
\end{table*}

\subsection{Barra-style return attribution}
\label{sec:barra}



Raw portfolio returns can flatter an agent that merely loads on a winning style: if high-volatility names rally during the window, a volatility-tilted portfolio scores high without any firm-level skill. We therefore decompose realized returns into market, style, and a residual that we treat as a closer proxy for stock-selection ability, while still reporting standard return/risk metrics in the headline table.

For each trading day \(t\) we estimate a weighted least-squares cross-sectional regression over the \(N_t\) investable stocks:
\vspace{-8pt}
\begin{equation}
\label{eq:barra}
r_{i,t}=f_{0,t}+\sum_{k=1}^{K} x_{i,k,t-1}\lambda_{k,t}+\varepsilon_{i,t},
\end{equation}
where \(r_{i,t}\) is stock \(i\)'s close-to-close return, \(f_{0,t}\) is a unit-exposure common-return component for the universe, \(x_{i,k,t-1}\) is the winsorized and cross-sectionally standardized exposure of stock \(i\) to style factor \(k\) (computed from information strictly before day \(t\)), \(\lambda_{k,t}\) is the estimated style-factor return, and \(\varepsilon_{i,t}\) is the stock-specific residual. Aggregating with the agent's pre-return weights \(w^p_{i,t-1}\) decomposes portfolio return as \(R^p_t = f_{0,t} + \sum_k (\sum_i w^p_{i,t-1}x_{i,k,t-1})\lambda_{k,t} + \alpha^p_t\), where the residual term \(\alpha^p_t = \sum_i w^p_{i,t-1}\varepsilon_{i,t}\) is the factor-adjusted selection alpha.

We use nine cross-sectional style factors spanning momentum, volatility, liquidity, reversal, skewness, and volume--price interaction, screened on a pre-evaluation calibration window by variance inflation factor \(\mathrm{VIF}_k = 1/(1-R_k^2)\) to reduce multicollinearity without using evaluation-period information.



We apply Eq.~\eqref{eq:barra} day-by-day to every completed trajectory and report seed-averaged cumulative contributions in Table~\ref{tab:barra-results}. Since the intercept in Eq.~\eqref{eq:barra} is a unit-exposure common return component rather than an externally observed market-index return, we denote it as Common. By construction, \(\text{Common}+\text{Style}+\text{Selection Alpha}=\text{Port}\). Per-LLM interpretation is deferred to Section~\ref{sec:disc-alpha}.


\begin{table}[t]
\centering
\small
\setlength{\tabcolsep}{4pt}
\begin{tabular}{lrrrr}
\toprule
Model & Com. & Sty. & Sel. $\alpha$ & Port \\
\midrule
Claude Opus 4.7     & $+35.0$ & $+16.6$ & $+0.2$  & $+51.8$ \\
Qwen3.6-Plus        & $+41.8$ & $+29.2$ & $-0.7$  & $+70.3$ \\
Doubao-Seed-2.0     & $+42.5$ & $+15.5$ & $-4.7$  & $+53.3$ \\
MiniMax-M2.7        & $+29.6$ & $+26.4$ & $-7.4$  & $+48.6$ \\
Step-3.5-Flash      & $+29.5$ & $+17.5$ & $-13.7$ & $+33.3$ \\
GPT-5.5             & $+40.0$ & $+11.8$ & $-14.9$ & $+36.9$ \\
Gemini-3.1-Pro      & $+38.3$ & $+23.6$ & $-38.5$ & $+23.4$ \\
DeepSeek-V4-Pro     & $+38.5$ & $+23.1$ & $-43.0$ & $+18.6$ \\
GLM-5.1             & $+39.5$ & $+27.5$ & $-61.9$ & $+5.1$  \\
Kimi-K2.6           & $+32.1$ & $+22.0$ & $-77.8$ & $-23.7$ \\
\bottomrule
\end{tabular}
\caption{Seed-averaged Barra attribution (\%) on the 2024--2026 window, sorted by selection alpha.}
\label{tab:barra-results}
\vspace{-15pt}
\end{table}

%% file: section/discussion.tex
\section{Discussion}
\label{sec:discussion}

\subsection{Masking reduces memory leakage}
\label{sec:disc-leakage}

Order rationales recorded by the anchor model on matched runs make the contamination channel directly visible. Under \textit{bright}, justifications interleave factor inputs with content the input does not carry---brand identity, sector taxonomy, and, most diagnostically, specific historical trading days (e.g.\ \emph{``the Oct-$18$ limit-up move''} for SMIC, a date unrecoverable from the price-only channel). Under \textit{blinded}, the same model receives identical numbers under anonymized aliases (\texttt{asset\_0006}, \texttt{asset\_0013}), and rationales reduce to pure factor ranks (\emph{``$20$-day $+27.4\%$, low vol $0.0255$''}); the narrative layer is removed while the factor layer is preserved.

This qualitative shift coincides with a measurable change in trading behavior. In \textit{memory-only}---where the prompt carries no factor data and the identifier is the only candidate signal---the anchor returns $-0.16\%$ under \textit{bright} and exactly $0.00\%$ under \textit{blinded} (Table~\ref{tab:mode-bright-blinded}): the ticker-visible model trades on identity priors and loses; the anonymized model holds cash. The disappearing trades are precisely those the rationale analysis attributes to pretraining content, and prompt-level restriction alone does not reproduce the ablation---data-side masking does work that no in-prompt instruction has been shown to replicate. Section~\ref{sec:abl-deanon} certifies the mask via an independent ten-attacker probe.

\subsection{Cumulative return overstates skill}
\label{sec:disc-alpha}

The Barra attribution in Table~\ref{tab:barra-results} concentrates almost all cross-agent variance in factor-adjusted selection alpha. Market contributions span a $13$pp range and style contributions a $17$pp range---every agent, including the worst performers, harvests $+11\%$ to $+29\%$ from systematic factor tilts---while selection alpha spans $+0.2\%$ to $-77.8\%$, a $78$pp range that dwarfs the other two combined. The implied ranking diverges sharply from the cumulative-return one: the total-return top four are Qwen3.6-Plus, GPT-5.5, Doubao-Seed-2.0, and Claude Opus 4.7, but on selection alpha only Claude Opus 4.7 ($+0.2\%$) is at par and nine of ten agents post negative selection alpha. The bottom four (Kimi-K2.6, GLM-5.1, DeepSeek-V4-Pro, Gemini-3.1-Pro) share the same broad factor tilts as the top group (Section~\ref{sec:barra-exposure}); their failure is not a stylistic miss but a within-factor selection failure---holding factor loading constant, their individual picks underperform what the loading alone would have delivered. A benchmark that uses cumulative return as its sole criterion therefore systematically over-credits agents that converge on the prevailing factor portfolio and under-credits within-factor selection skill.

\subsection{Factor Exposure Preferences}
\label{sec:barra-exposure}
For every model we compute the portfolio-weighted, $z$-scored exposure $\sum_i \omega_{i,t}\, x_{i,k,t-1}$ to each style factor in \ref{sec:barra}, average it across invested days in 2024-01-02--2026-04-10, then average within each cohort. The full nine-factor breakdown is in Appendix~\ref{sec:appendix-exposure} (Table~\ref{tab:exposure-comparison}); the two cohorts agree on three universal signals---long-horizon momentum, the $52$-week-high anchor, and an aversion to illiquid names---and occupy near-mirror regimes on the remaining five. ML models form a classical \emph{low-volatility momentum book}: net short realized volatility, amount variability, and intraday momentum, with $15$ of $18$ sharing the same sign across all nine factors. LLMs flip those tilts, with intraday momentum ($+1.06\sigma$) as their largest single loading and positive exposures on realized volatility and volume--return correlation. In one line: ML books hold quiet trending large caps; LLM books load on high-volatility, recently moving names.

%% file: section/conclusion.tex
\section{Conclusion}
\label{sec:conclusion}

\benchname{} closes two gaps in how end-to-end LLM trading is evaluated. On the input side, a four-level mask anonymizes tickers, calendar dates, and tool-return timestamps consistently across prompts and tools, with an external attacker audit certifying that the channel is closed. On the output side, a Barra-style cross-sectional regression decomposes each agent's daily return into market, style, and stock-selection alpha, separating skill from passive factor harvesting. Across ten frontier LLM agents on a 2024--2026 window against eighteen Alpha9-trained baselines, masking changes the shape of trading---rationales drop from brand narratives to factor ranks---and the cumulative returns a leaderboard would report come from passive market and style exposure rather than from stock-selection alpha. The same pair---a data-side mask that closes the memory channel and an attribution layer that exposes factor luck---transfers directly to news-driven and fundamentals-based LLM evaluation, and we release \benchname{} as a template for that broader benchmark family.

%% file: section/limitations.tex
\section*{Limitations}
\label{sec:limitations}

The current implementation focuses on the CSI300 A-share universe with a price-only observation channel (open, high, low, close, volume, and derived technical factors); it does not cover US or Hong Kong equities, nor multimodal inputs such as archived news, filings, or macroeconomic series. These dimensions could affect agent behaviour but are out of scope for the current framework; we plan to extend market coverage and add a timestamp-aligned multimodal track in future iterations. The full mask $\times$ mode $\times$ window grid is populated only on the anchor model Step-3.5-Flash, while the other nine frontier LLMs are evaluated solely under the core \textit{blinded $\times$ open-research} condition; running the full grid on all ten models is compute-bound rather than methodologically blocked, as each LLM on the long window with three seeds takes 24--72 hours of vendor-API wall-clock. We have additionally conducted fragmentary local runs on further models (e.g., GPT, Opus and a Doubao variant), and a multi-model depth replication would yield more granular comparisons across the design axes.

%% file: section/appendix.tex
\section{Rationale case study across the four mask conditions}
\label{sec:appendix-cases}

The motivating observation in Section~\ref{sec:intro} is that the same model, under price-only observation, writes qualitatively different rationales depending on what identifying handles are visible. The four masking conditions in Section~\ref{sec:masking} predict a specific decomposition: \textit{stock-blind} should erase ticker-anchored priors (brand, sector, business model) while leaving any calendar-anchored prior intact; \textit{date-blind} should erase calendar-anchored priors while leaving ticker-anchored ones intact; \textit{blinded} should erase both; \textit{bright} should permit both. Boxes below reproduce verbatim rationales for the same anchor model under each condition on equivalent numeric inputs.

\begin{tcolorbox}[
  float=h, floatplacement=h, label={box:bright-cases},
  enhanced, breakable,
  colback=brightbg, colframe=brightrule,
  boxrule=0.6pt, arc=2pt,
  fonttitle=\bfseries\footnotesize\sffamily,
  coltitle=white, colbacktitle=brightrule,
  left=4pt, right=4pt, top=4pt, bottom=4pt,
  title={\textsc{bright} \quad real ticker IDs visible \quad (4 verbatim rationales)},
]
\footnotesize
\textbf{[SH600519 BUY]} ``\textbf{Kweichow Moutai}, \emph{liquor leader}; low volatility ($0.0234$), drawdown well controlled ($-7.3\%$), 5-day return $+15.2\%$ ranks among the top---suitable as a steady allocation.''

\medskip
\textbf{[SH600900 BUY]} ``\textbf{Yangtze Power}, low volatility $0.01$, 5-day $+3.1\%$, 20-day $+6.8\%$; in a weak market, add a new $5\%$ position as a \emph{defensive utility} core holding.''

\medskip
\textbf{[SH601939 BUY]} ``\textbf{CCB} 20-day momentum $+11\%$ tops the screen, low volatility, shallow drawdown---fits a \emph{large-cap low-vol high-momentum} style.''

\medskip
\textbf{[SH688981 BUY]} ``\textbf{Semiconductor manufacturing leader}, beneficiary of an \emph{industry-recovery expectation}; 20-day return $+108\%$, the Oct-18 limit-up move shows strong drive, volatility $8.31\%$ relatively contained---suitable for allocation.''
\end{tcolorbox}

\begin{tcolorbox}[
  float=h, floatplacement=h, label={box:stock-blind-cases},
  enhanced, breakable,
  colback=blindedbg!40!brightbg, colframe=brightrule!60!blindedrule,
  boxrule=0.6pt, arc=2pt,
  fonttitle=\bfseries\footnotesize\sffamily,
  coltitle=white, colbacktitle=brightrule!60!blindedrule,
  left=4pt, right=4pt, top=4pt, bottom=4pt,
  title={\textsc{stock-blind} \quad anonymized ticker, real date \quad (3 verbatim rationales)},
]
\footnotesize
\textbf{[asset\_0004 SELL]} ``asset\_0004 current weight $10.38\%$ exceeds the $10\%$ single-name cap, must reduce $160$ shares to comply; the name has $20$-day return $+23.7\%$ and a strong showing, but \emph{valuation is now elevated}---take partial profit, retain the core.''

\medskip
\textbf{[asset\_0017 BUY]} ``$20$-day return tops the screen at $+33.6\%$, long-run trend is clear; despite weaker short-term momentum the \emph{fundamental support remains strong}.''

\medskip
\textbf{[asset\_0023 BUY]} ``$5$-day return $+12.9\%$ ranks \#1, short-term momentum strong; despite a sharper $20$-day drawdown, the \emph{current valuation is reasonable}---allocate moderately to catch the trend.''
\end{tcolorbox}

\begin{tcolorbox}[
  float=h, floatplacement=h, label={box:date-blind-cases},
  enhanced, breakable,
  colback=brightbg!40!blindedbg, colframe=blindedrule!60!brightrule,
  boxrule=0.6pt, arc=2pt,
  fonttitle=\bfseries\footnotesize\sffamily,
  coltitle=white, colbacktitle=blindedrule!60!brightrule,
  left=4pt, right=4pt, top=4pt, bottom=4pt,
  title={\textsc{date-blind} \quad real ticker, anonymized date \quad (3 verbatim rationales)},
]
\footnotesize
\textbf{[SH600111 BUY]} ``\textbf{SH600111} has low volatility and \emph{strong defensive properties}; the \emph{rare-earth sector} carries allocation value, helps diversify portfolio risk.''

\medskip
\textbf{[SH601857 BUY]} ``\textbf{SH601857} $20$-day return $+21.4\%$, low volatility $0.0211$, drawdown controlled at $-5.45\%$; \emph{low valuation, high safety margin}.''

\medskip
\textbf{[SH600362 BUY]} ``\textbf{SH600362} short-term momentum strong and volatility low; $20$-day return $+27.6\%$ with controlled drawdown---replace some \emph{high-valuation positions}.''
\end{tcolorbox}

\begin{tcolorbox}[
  float=h, floatplacement=h, label={box:blinded-cases},
  enhanced, breakable,
  colback=blindedbg, colframe=blindedrule,
  boxrule=0.6pt, arc=2pt,
  fonttitle=\bfseries\footnotesize\sffamily,
  coltitle=white, colbacktitle=blindedrule,
  left=4pt, right=4pt, top=4pt, bottom=4pt,
  title={\textsc{blinded} \quad anonymized aliases \quad (4 verbatim rationales)},
]
\footnotesize
\textbf{[asset\_0006 BUY]} ``asset\_0006 has 20-day return $+27.4\%$, low volatility $0.0255$, and small drawdown $-3.8\%$; risk/return ratio excellent, suitable to add up to the $10\%$ cap.''

\medskip
\textbf{[asset\_0013 BUY]} ``asset\_0013 5-day return $+14.5\%$ ranks \#2, 20-day $+27.6\%$ ranks \#1, low volatility and a maximum drawdown of only $-3.87\%$; risk/return ratio superior, build a new $8\%$ position.''

\medskip
\textbf{[asset\_0028 SELL]} ``asset\_0028 20-day return only $+14.4\%$ ranks low and the unrealised loss is largest; risk/return unfavourable, close the position to release capital.''

\medskip
\textbf{[asset\_0011 SELL]} ``asset\_0011 has the largest unrealised loss and is not in the top-15 by 20-day return; underperforms the market average, replace with a stronger candidate.''
\end{tcolorbox}

The four boxes localize the gap predicted at the top of the section. Under \textit{bright} the rationales name brand, sector, and macro identities (\textsc{Kweichow Moutai}, \textsc{Yangtze Power}, \textsc{CCB}, semiconductor industry recovery) that OHLCV cannot supply; under \textit{blinded} the same model on identical numbers produces only ranks, momentum, volatility, drawdown, and unrealised PnL. The intermediate two boxes show why: \textit{date-blind} retains brand and sector priors in full force (\texttt{SH600111} ``rare-earth sector ... defensive''; \texttt{SH601857} ``low valuation, high safety margin''), while \textit{stock-blind} reduces them to generic valuation filler with no anchoring referent. The ticker handle, not the calendar handle, carries the dominant pretraining prior.

\section{Tool interface specification}
\label{sec:appendix-tools}

The open-research decision mode exposes six read-only tools (Table~\ref{tab:tools}). Tool arguments are un-masked by the harness before querying the data layer; the returned payload is re-masked before being handed back to the model, so an agent in \textit{blinded} mode never observes a real ticker or real calendar date even transiently. All tools return JSON dictionaries with explicit type schemas.

\begin{table}[h]
\centering
\footnotesize
\setlength{\tabcolsep}{4pt}
\begin{tabular}{@{}llp{2.3cm}@{}}
\toprule
Tool & Key args & Returned fields \\
\midrule
\texttt{get\_market\_context}  & ---             & indices, calendar, sector tone \\
\texttt{screen\_candidates}    & factors, $k$    & ranked aliases + factor values \\
\texttt{get\_stock\_snapshot}  & alias, lookback & OHLCV bars, factors, prev close \\
\texttt{compare\_candidates}   & aliases, dims   & side-by-side metric grid \\
\texttt{portfolio\_state}      & ---             & holdings, cash, T+1 locks \\
\texttt{risk\_check}           & target weights  & violations, projected limits \\
\bottomrule
\end{tabular}
\caption{Six read-only tools under \textit{open-research}. The agent may invoke any tool any number of times within a step.}
\label{tab:tools}
\end{table}

\clearpage
\onecolumn
\section{Prompts and a worked masked trace}
\label{sec:appendix-prompts}

One system prompt is given per decision mode in Section~\ref{sec:modes}, and a single user-message template is shared across modes.

\subsection*{System prompt: \textit{open-research} (headline configuration)}

\begin{lstlisting}[language=]
You are an autonomous Chinese A-share trading agent. Within the
current backtest trading day, conduct research, judgement, order
submission, and self-constraint based on available data; emit a
single valid JSON trading decision.

## Goals
- Maximise the overall benchmark score within the hard constraints.
- Decide on your own whether to trade, how many names, concentrated
  or diversified, and whether to use `target_weight` or `shares`.
- If no identifiable edge exists, holding cash or maintaining the
  status quo is an acceptable decision.

## Degrees of freedom
- No fixed workflow is required; decide which tools to call, in what
  order and frequency, and when to stop research and submit.

## Information boundary
- Decisions may rely only on the current message, system context,
  and tool returns; do not assume any externally available
  information.
- All tools have time semantics strictly earlier than the current
  backtest day: market-data tools see the previous trading day and
  earlier; news / report / announcement / search tools have their
  returns filtered as of the current day, but you should still
  verify the date fields in returns to avoid indirectly referencing
  future information.
- The current day's open, close, and intraday data are not visible.

## Hard constraints
1. Long-only; no shorting.
2. Only stocks in today's tradable universe may be operated on.
3. A-share T+1: a stock bought today can only be sold the next day.
4. Each order must contain `stock_id`, `side`, `confidence`, and
   `reason`, with `reason` of at least 10 characters.
5. If no action is intended, return an empty `orders` list.

## Autonomous decision space
- Neither concentration nor the number of holdings has a hard upper
  bound; you may concentrate heavily (e.g., 1-3 names) or diversify
  broadly. Weigh win rate, risk budget, and candidate quality on
  your own.
- State your portfolio-construction rationale in `overall_reason`.

## Limit-board rules (engine layer)
- Price limits are board-differentiated: main / SME (6xxxxx / 00xxxx)
  +/-9.5%; ChiNext (30xxxx) and STAR (688xxx) +/-19.5%; Beijing
  Stock Exchange (8xxxxx / 4xxxxx) +/-29.5%.
- A next-day open at the upper limit rejects BUY; lower limit
  rejects SELL; one-sided days (open == high == low at a limit
  price) are unfillable.
- Tools expose `board / limit_pct / limit_up / limit_down` and
  limit-hit warnings per stock.

## Execution details
- Orders execute sequentially in the order given; on days with both
  buys and sells, list SELL before BUY to avoid running out of cash.
- `target_weight` is the final desired weight; do not re-submit BUY
  for a position whose current weight is already near or above it.
- When opening longs, retain a 1%-3% cash buffer (fees, lot rounding,
  spread).
- SELL only against holdings with `shares_available > 0`.
- When cash or sellable shares are uncertain, prefer to under-trade,
  lower the target, or do nothing.

## Decision requirements
- Emit an executable trading decision, not an analytical report.
- Order intentions must be clear, non-conflicting, and free of
  meaningless instructions.
- `confidence` in [0, 1]; `overall_reason` >= 20 characters.

## Output format
Emit a single JSON object only, starting with `{` and ending with
`}`; no markdown code-block wrappers, no extra text, no comments,
no tool-call summaries. See Appendix for the full action schema.
\end{lstlisting}

\subsection*{System prompt: \textit{fixed-candidate}}

\begin{lstlisting}[language=]
You are a professional Chinese A-share quantitative analyst. The
current experimental mode is fixed-candidate: a fixed candidate
pool is provided; you cannot call tools and cannot extend the
candidate set on your own.

Important notes
- Stock identifiers appear in the form provided in the input
  (e.g., asset_0042); use the input identifiers verbatim.
- The current step identifier may be a real date or a relative
  index. Do not use external memory, news, real-stock common
  knowledge, or any data not provided.
- BUY only from the fixed candidate pool; SELL only against current
  holdings; an empty `orders` list is acceptable.
- Long-only; A-share T+1 (sell the next day at the earliest);
  concentration and number of holdings have no hard upper bound,
  decide on your own.
- Each order must contain `stock_id`, `side`, `confidence`, and
  `reason`; `target_weight` and `shares` are mutually exclusive.

The output must be a single pure JSON object (no markdown, no
extra text).
\end{lstlisting}

\subsection*{System prompt: \textit{memory-only}}

\begin{lstlisting}[language=]
You are a professional Chinese A-share quantitative analyst. The
current experimental mode is memory-only: the system provides only
today's tradable universe (identifiers), your account state, and
constraints. No price, volume, return, volatility, factor, or
indicator data is provided, and no tool calls are permitted.

Important notes
- The input contains only stock identifiers. An identifier may be
  a real ticker (e.g., SH600519) or an anonymous alias (e.g.,
  asset_0042); use the input identifiers verbatim.
- The current step identifier may be a real date or a relative
  index.
- Your decision may rely only on the identifiers, account state,
  and constraints in the input. If you have a view on a given
  identifier, you may act on it; if you have no basis for a
  decision, you may return an empty `orders` list.
- BUY only from today's tradable universe (identifiers present in
  the input); SELL only against current holdings; no action is
  always allowed.
- Long-only; A-share T+1; concentration and number of holdings have
  no hard upper bound.
- Each order must contain `stock_id`, `side`, `confidence`, and
  `reason`; `target_weight` and `shares` are mutually exclusive.

The output must be a single pure JSON object (no markdown, no
extra text).
\end{lstlisting}

\subsection*{User-message template}

A single user-message template is shared across decision modes; the
\texttt{\{...\}} placeholders are filled at runtime by the harness.
Under \textit{blinded}, \texttt{date\_label} renders as a relative
day index (e.g., \texttt{day\_+12}) and \texttt{positions\_table} /
\texttt{universe\_preview} / \texttt{features\_table} contain only
\texttt{asset\_NNNN} aliases.

\begin{lstlisting}[language=]
## Backtest step: {date_label}

## Constraint summary
- Concentration and number of holdings have no hard upper bound;
  decide on your own.
- Short selling: {allow_short}
- T+1 settlement: yes

## Current holdings and account
- Total value: {total_value} CNY
- Cash: {cash} CNY ({cash_weight} of NAV)
- {num_positions} positions:
{positions_table}

## Today's tradable universe ({universe_size} names)
{universe_preview}

## Market summary over the last {obs_window} observations
## ({features_count} names; data cutoff T-1 close)
{features_table}

## Previous-step execution outcome
{prev_execution_summary}

## Portfolio performance
- Cumulative return since the start of the backtest:
  {portfolio_ret_cum}
- CSI300 cumulative return over the same window:
  {benchmark_ret_cum}

Emit today's trading decision based on the above (pure JSON, no
markdown).
\end{lstlisting}

\subsection*{Worked masked trace: one step under \textit{blinded $\times$ open-research}}
\label{sec:appendix-trace}

Box~\ref{box:trace} reproduces the agent's view of a single step under \textit{blinded $\times$ open-research} (anchor model Step-3.5-Flash, window W1, step 0). The agent issues four tool calls (\texttt{get\_market\_context}, \texttt{screen\_candidates}, \texttt{compare\_candidates}, \texttt{risk\_check}) and one submission; we show three of the calls and the final action. Every identifier the agent sees is an alias and every date is a relative day index; the executor un-masks aliases back to real tickers post-hoc for order execution, but the agent itself never observes a real identifier in its inputs or outputs.

\begin{tcolorbox}[
  float=h, floatplacement=h, label={box:trace},
  enhanced, breakable,
  colback=blindedbg, colframe=blindedrule,
  boxrule=0.6pt, arc=2pt,
  fonttitle=\bfseries\footnotesize\sffamily,
  coltitle=white, colbacktitle=blindedrule,
  left=4pt, right=4pt, top=4pt, bottom=4pt,
  title={\textsc{blinded $\times$ open-research} \quad anchor model, W1 step 0 \quad agent's view},
]
\footnotesize\ttfamily
\textbf{[1] tool\_call} \quad get\_market\_context()

\smallskip
\textbf{[1] tool\_result} \quad \{as\_of\_date: \textit{day\_+0}, data\_cutoff: \textit{day\_-3}, contains\_current\_day\_market\_data: false, market: csi300, universe\_size: 300, ...\}

\medskip
\textbf{[2] tool\_call} \quad screen\_candidates(sort\_by=ret\_20d, top\_k=15)

\smallskip
\textbf{[2] tool\_result} \quad \{as\_of\_date: \textit{day\_+0}, data\_cutoff: \textit{day\_-3}, candidates: [\\
\hspace*{1em}\{stock\_id: \textbf{asset\_0001}, prev\_close: 3.034, ret\_1d: 0.0185, ret\_5d: 0.1123, ret\_20d: 0.3355, vol\_20d: 0.0302, drawdown\_20d: $-0.055$, rank: 1\},\\
\hspace*{1em}\{stock\_id: \textbf{asset\_0002}, prev\_close: 0.567, ret\_1d: 0.0996, ret\_5d: 0.2213, ret\_20d: 0.3104, vol\_20d: 0.0272, drawdown\_20d: $-0.045$, rank: 2\}, ...\,\}

\medskip
\textbf{[3] tool\_call} \quad risk\_check(draft\_orders=[\\
\hspace*{1em}\{stock\_id: \textbf{asset\_0001}, side: BUY, target\_weight: 0.10, confidence: 0.85, reason: ``\textit{20-day return ranks \#1 (+33.6\%); short-term momentum strong; core allocation}''\},\\
\hspace*{1em}\{stock\_id: \textbf{asset\_0002}, side: BUY, target\_weight: 0.10, confidence: 0.82, reason: ``\textit{1-day +9.96\% and 5-day +22.1\% both rank \#1; explosive short-term strength}''\}, ...\,])

\smallskip
\textbf{[3] tool\_result} \quad \{valid: true, violations: [], projected\_weights: \{asset\_0001: 0.10, asset\_0002: 0.10, ...\}\}

\medskip
\textbf{[final] submit\_action} \quad \{orders: [\\
\hspace*{1em}\{stock\_id: \textbf{asset\_0001}, side: BUY, target\_weight: 0.10, confidence: 0.85, reason: ``\textit{...}''\},\\
\hspace*{1em}\{stock\_id: \textbf{asset\_0002}, side: BUY, target\_weight: 0.10, confidence: 0.82, reason: ``\textit{...}''\}, ...\,],\\
overall\_reason: ``\textit{High-momentum, low-volatility names dominate; build five-name core positions and retain cash buffer.}''\}
\end{tcolorbox}

The trace is the empirical realization of the masking-consistency claim in Section~\ref{sec:masking}: aliases and relative day indices propagate end-to-end through every tool argument, tool return, and the agent's submitted action. Field-level masking is also visible: \texttt{as\_of\_date} and \texttt{data\_cutoff} are relative indices, candidate \texttt{stock\_id}s are aliases, and the \texttt{reason} strings reference only numeric features (ranks, returns, volatility) rather than brand or sector language.

\clearpage
\twocolumn
\section{Action schema}
\label{sec:appendix-action}

The submission phase emits one JSON object per step, schema-validated by the harness:

\begin{lstlisting}[language=]
{
  "orders": [
    { "stock_id":      "asset_0042",
      "side":          "BUY" | "SELL",
      "target_weight": 0.05,
      "confidence":    0.75,
      "reason":        "..." }
  ],
  "overall_reason": "..."
}
\end{lstlisting}

The \texttt{confidence} field feeds the calibration metrics (Appendix~\ref{sec:appendix-metrics}); the \texttt{reason} field feeds the motivation analysis (Section~\ref{sec:intro}) and the case study in Appendix~\ref{sec:appendix-cases}.

\section{Style factors}
\label{sec: style factors}
We select nine classic factors with low mutual correlation via VIF, as is shown in Table \ref{tab:barra-factors}
\begin{table}[h]
\centering
\small
\begin{tabular}{ll}
\toprule
Factor & Description \\
\midrule
\textsc{MOM\_12\_1} & 12-minus-1 month momentum \\
\textsc{RV\_60}     & 60-day realized volatility \\
\textsc{ILLIQ}      & Amihud illiquidity \\
\textsc{REV\_ON}    & Overnight return \\
\textsc{MOM\_ID}    & Intraday momentum \\
\textsc{SKEW}       & Negative return skewness \\
\textsc{CORR\_PV}   & Volume--return correlation \\
\textsc{HIGH\_52W}  & 52-week high anchor \\
\textsc{CV\_VOL}    & Amount coefficient of variation \\
\bottomrule
\end{tabular}
\caption{Nine style factors used in the cross-sectional attribution model. }
\label{tab:barra-factors}
\end{table}

\section{Cohort-level factor exposures}
\label{sec:appendix-exposure}

Table~\ref{tab:exposure-comparison} reports the portfolio-weighted, $z$-scored exposures from Section~\ref{sec:barra-exposure}.

\begin{table}[!htbp]
\centering
\small
\begin{tabular}{lccc}
\toprule
Factor & ML  & LLM  & ML $-$ LLM \\
\midrule
\textsc{MOM\_12\_1} & $+0.34$ & $+0.40$ & $-0.07$ \\
\textsc{RV\_60}     & $-0.66$ & $+0.23$ & $\mathbf{-0.89}$ \\
\textsc{ILLIQ}      & $-0.34$ & $-0.32$ & $-0.02$ \\
\textsc{REV\_ON}    & $+0.12$ & $+0.28$ & $-0.17$ \\
\textsc{MOM\_ID}    & $-0.12$ & $+1.06$ & $\mathbf{-1.17}$ \\
\textsc{SKEW}       & $+0.42$ & $-0.08$ & $\mathbf{+0.50}$ \\
\textsc{CORR\_PV}   & $-0.26$ & $+0.33$ & $\mathbf{-0.59}$ \\
\textsc{HIGH\_52W}  & $+0.80$ & $+0.77$ & $+0.03$ \\
\textsc{CV\_VOL}    & $-0.60$ & $+0.11$ & $\mathbf{-0.71}$ \\
\bottomrule
\end{tabular}
\caption{Mean portfolio-weighted $z$-scored factor exposures by cohort; positive values indicate net long exposure, negative values net short. Bold entries mark $|\text{gap}|>0.4\sigma$.}
\label{tab:exposure-comparison}
\end{table}

\section{Ten-dimensional metric definitions}
\label{sec:appendix-metrics}

\paragraph{Returns and risk.}
\textit{Total return} is the geometric return over the evaluation period from end-of-day NAV. \textit{Sharpe ratio} is annualized with $r_f = 0$ and a 252-day convention. \textit{Maximum drawdown} is the largest peak-to-trough decline in NAV. \textit{Information ratio} is the annualized mean of daily excess return over CSI300 divided by its annualized standard deviation.

\paragraph{Behavior.}
\textit{Annualized turnover} is the sum of absolute weight changes per step, scaled to a 252-day year. \textit{HHI} is the average across steps of $\sum_i w_i^2$ over current holding weights. \textit{Cash utilization} is the average fraction of NAV held as cash. \textit{Abstention rate} is the fraction of steps producing zero orders after retries.

\paragraph{Reliability and calibration.}
\textit{Parse-failure rate} is the fraction of action emissions failing JSON schema validation after three retries. \textit{Tool-validity rate} is the fraction of tool calls whose arguments parse and dispatch successfully. \textit{ECE} bins per-order \texttt{confidence} into 10 equal-width bins and reports the weighted mean absolute gap between bin confidence and bin accuracy. \textit{Brier score} treats per-order confidence as the probability that the order's realized next-period return is positive.

\section{Qlib factor model configuration}
\label{sec:appendix-qlib}

The eighteen Qlib factor baselines use the published Qlib defaults for the CSI300 universe (hyperparameters unchanged from the official model-zoo configuration files); only the ranking signal differs across models. The portfolio constructor is identical across models: top-20 by predicted score with a cost-aware rebalancing threshold of $0.06$. Table~\ref{tab:qlib-full} reports the complete leaderboard for all eighteen baselines; the main paper (Table~\ref{tab:ten-llm}) shows eight representatives.

\paragraph{A-share price-limit thresholds (used by both LLM agents and Qlib baselines).}
The execution engine applies a small buffer below each board's official daily limit to avoid fills at unreachable prices: $9.5\%$ for main-board issues (Shanghai \texttt{6xxxxx} and Shenzhen \texttt{00xxxx}, daily limit $\pm 10\%$), $19.5\%$ for ChiNext (\texttt{30xxxx}) and STAR Market (\texttt{688xxx}, both $\pm 20\%$), and $29.5\%$ for the Beijing Stock Exchange (\texttt{8xxxxx}, $\pm 30\%$).

\begin{table*}[!htbp]
\centering
\footnotesize
\setlength{\tabcolsep}{4pt}
\begin{tabular}{lrrrrrr}
\toprule
Model & Total ret. & Excess & Sharpe & MDD & IR & Turn. \\
\midrule
\texttt{SFM}            & $+86.58\%$ & $+49.66\%$ & $2.02$ & $-7.41\%$  & $+0.57$ & $22.47$  \\
\texttt{DoubleEnsemble} & $+73.66\%$ & $+36.74\%$ & $1.41$ & $-18.24\%$ & $+0.41$ & $72.84$  \\
\texttt{CatBoost}       & $+72.97\%$ & $+36.05\%$ & $1.43$ & $-16.74\%$ & $+0.40$ & $29.96$  \\
\texttt{XGBoost}        & $+66.41\%$ & $+29.49\%$ & $1.12$ & $-17.75\%$ & $+0.31$ & $101.66$ \\
\texttt{LightGBM}       & $+66.22\%$ & $+29.30\%$ & $1.29$ & $-15.59\%$ & $+0.33$ & $46.18$  \\
\texttt{ADARNN}         & $+66.13\%$ & $+29.21\%$ & $1.66$ & $-8.52\%$  & $+0.35$ & $16.19$  \\
\texttt{KRNN}           & $+62.09\%$ & $+25.17\%$ & $1.63$ & $-8.88\%$  & $+0.31$ & $20.80$  \\
\texttt{Transformer}    & $+60.35\%$ & $+23.43\%$ & $1.61$ & $-11.29\%$ & $+0.27$ & $11.22$  \\
\texttt{Sandwich}       & $+56.57\%$ & $+19.65\%$ & $1.49$ & $-8.90\%$  & $+0.24$ & $29.78$  \\
\texttt{ADD}            & $+54.03\%$ & $+17.11\%$ & $1.47$ & $-9.92\%$  & $+0.20$ & $22.52$  \\
\texttt{Localformer}    & $+53.02\%$ & $+16.10\%$ & $1.41$ & $-10.59\%$ & $+0.19$ & $13.32$  \\
\texttt{TCN}            & $+52.33\%$ & $+15.41\%$ & $1.31$ & $-10.27\%$ & $+0.18$ & $30.17$  \\
\texttt{TabNet}         & $+52.25\%$ & $+15.33\%$ & $1.38$ & $-10.61\%$ & $+0.18$ & $26.13$  \\
\texttt{LSTM}           & $+43.92\%$ & $+7.00\%$  & $1.12$ & $-11.76\%$ & $+0.06$ & $78.03$  \\
\rowcolor{gray!18}
CSI300 buy-and-hold     & $+36.92\%$ & $0.00\%$   & ---    & ---        & ---     & $0.00$   \\
\texttt{MLP}            & $+34.08\%$ & $-2.84\%$  & $0.69$ & $-21.75\%$ & $-0.01$ & $18.48$  \\
\texttt{GRU}            & $+34.05\%$ & $-2.87\%$  & $0.92$ & $-13.50\%$ & $-0.07$ & $57.79$  \\
\texttt{Linear}         & $+29.60\%$ & $-7.32\%$  & $0.74$ & $-13.88\%$ & $-0.12$ & $15.61$  \\
\texttt{ALSTM}          & $+22.67\%$ & $-14.25\%$ & $0.64$ & $-15.15\%$ & $-0.22$ & $95.31$  \\
\bottomrule
\end{tabular}
\caption{Complete leaderboard for all eighteen trained Qlib factor models on the long 2024-01-01 to 2026-04-10 window, sorted by total return. CSI300 buy-and-hold inserted at its rank position.}
\label{tab:qlib-full}
\end{table*}

\section{Harness configuration}
\label{sec:appendix-config}

Configuration values used in the headline ten-LLM evaluation (\textit{blinded $\times$ open-research}, long window 2024-01-01 to 2026-04-10).

\begin{table}[!htbp]
\centering
\footnotesize
\setlength{\tabcolsep}{4pt}
\begin{tabular}{@{}llr@{}}
\toprule
Group & Parameter & Value \\
\midrule
\multirow{8}{*}{Execution}
  & benchmark           & CSI300 \\
  & initial cash (CNY)  & $1{,}000{,}000$ \\
  & buy cost (bps)      & $5$ \\
  & sell cost (bps)     & $15$ \\
  & min cost (CNY)      & $5$ \\
  & deal price          & next-day open \\
  & T+1 settlement      & yes \\
  & long-only           & yes \\
\midrule
\multirow{8}{*}{Agent}
  & max candidates / step  & $100$ \\
  & max tool calls / step  & $99$ \\
  & max positions held     & $300$ \\
  & max single weight      & $1.00$ \\
  & schema retries         & $2$ \\
  & retry with feedback    & yes \\
  & min reason length      & $10$ \\
  & fallback action        & hold \\
\midrule
\multirow{4}{*}{LLM call}
  & temperature         & $0.0$ \\
  & max tokens          & $32{,}768$ \\
  & timeout (s)         & $600$ \\
  & inter-call gap (s)  & $3$ \\
\bottomrule
\end{tabular}
\caption{Headline-experiment harness configuration.}
\label{tab:harness-config}
\end{table}

\section{Mask sensitivity on additional frontier LLMs}
\label{sec:appendix-multimodel}

We replicate the \textit{bright} vs.\ \textit{blinded} contrast under \textit{open-research} on three additional frontier LLMs over ten regime-stratified windows with three seeds per cell; each cell reports the seed- and window-median (Table~\ref{tab:multimodel-mask}).

\begin{table}[!htbp]
\centering
\footnotesize
\setlength{\tabcolsep}{4pt}
\begin{tabular}{@{}llrrr@{}}
\toprule
Model & Mask & Ret.\ (\%) & Abst.\ & Turn. \\
\midrule
\multirow{2}{*}{\texttt{gpt-5.5}} & bright   & $-0.15$ & $0.18$ & $26.7$ \\
                                  & blinded  & $-1.43$ & $0.12$ & $29.9$ \\
\midrule
\multirow{2}{*}{\texttt{doubao-seed-2.0}} & bright   & $-2.97$ & $0.03$ & $31.3$ \\
                                          & blinded  & $-4.07$ & $0.00$ & $33.7$ \\
\midrule
\multirow{2}{*}{\texttt{claude-opus-4-7}} & bright   & $-0.40$ & $0.20$ & $21.4$ \\
                                          & blinded  & $-1.44$ & $0.02$ & $73.5$ \\
\bottomrule
\end{tabular}
\caption{Mask sensitivity across three frontier LLMs under \textit{open-research} on ten regime-stratified windows with three seeds per cell. \texttt{Ret.} is median total return; \texttt{Abst.} is median abstention rate; \texttt{Turn.} is median annualized turnover.}
\label{tab:multimodel-mask}
\end{table}

\section{De-anonymization probe details}
\label{sec:appendix-deanon}

\paragraph{Protocol.}
For each (stock, date) probe the harness calls \texttt{get\_market\_context}, \texttt{screen\_candidates}, and \texttt{get\_stock\_snapshot} with bars through the alias map with \textit{blinded} on, then saves the masked payload alongside the gold (ticker, date, sector, board). Cells receive Wilson 95\% confidence intervals.

\paragraph{Full-attacker breakdown.}
Table~\ref{tab:deanon-full} reports per-attacker recovery rates under \texttt{full}. Each attacker is given the same 200 randomly sampled probes; rows report rates over successful API responses only, excluding calls that failed to return a parseable answer. Top-1 ticker accuracy never exceeds $3.0\%$ against a random baseline of $0.3\%$, and joint success (ticker in top-5 \emph{and} date within $\pm 7$ trading days) never exceeds $1.5\%$. Board accuracy is intentionally high because the board field is not anonymized---the agent needs it for limit-price compliance---and serves as a sanity check.

\begin{table*}[!t]
\centering
\footnotesize
\setlength{\tabcolsep}{6pt}
\begin{tabular}{@{}lrrrrrrr@{}}
\toprule
Attacker & tk@1 & tk@5 & board & date $\pm 7$d & date $\pm 30$d & date $\pm 90$d & success \\
\midrule
\texttt{step-3.5-flash}    & $0.0\%$ & $2.8\%$  & $100.0\%$ & $4.4\%$  & $13.3\%$ & $26.7\%$ & $0.0\%$ \\
\texttt{claude-opus-4-7}   & $3.0\%$ & $10.2\%$ & $100.0\%$ & $1.0\%$  & $10.2\%$ & $22.8\%$ & $0.0\%$ \\
\texttt{doubao-seed-2.0}   & $1.5\%$ & $9.0\%$  & $100.0\%$ & $7.0\%$  & $14.0\%$ & $23.5\%$ & $0.0\%$ \\
\texttt{gpt-5.5}           & $2.0\%$ & $8.0\%$  & $\phantom{0}99.5\%$ & $7.5\%$  & $13.0\%$ & $20.5\%$ & $0.5\%$ \\
\texttt{qwen3.6-plus}      & $1.0\%$ & $4.6\%$  & $100.0\%$ & $6.2\%$  & $12.3\%$ & $28.7\%$ & $0.0\%$ \\
\texttt{minimax-m2.7}      & $1.1\%$ & $4.9\%$  & $\phantom{0}99.5\%$ & $3.3\%$  & $12.0\%$ & $21.7\%$ & $0.0\%$ \\
\texttt{deepseek-v4-pro}   & $0.5\%$ & $5.5\%$  & $100.0\%$ & $10.5\%$ & $18.0\%$ & $34.0\%$ & $0.5\%$ \\
\texttt{glm-5.1}           & $0.0\%$ & $6.4\%$  & $100.0\%$ & $16.0\%$ & $24.5\%$ & $45.7\%$ & $1.1\%$ \\
\texttt{kimi-k2.6}         & $2.2\%$ & $9.8\%$  & $100.0\%$ & $1.6\%$  & $11.5\%$ & $29.0\%$ & $0.5\%$ \\
\texttt{gemini-3.1-pro}    & $3.0\%$ & $8.6\%$  & $100.0\%$ & $14.7\%$ & $21.8\%$ & $36.5\%$ & $1.5\%$ \\
\midrule
random baseline           & $0.3\%$ & $1.4\%$  & $25.0\%$  & $1.9\%$  & $\phantom{0}8.2\%$ & $24.7\%$ & $0.0\%$ \\
\bottomrule
\end{tabular}
\caption{De-anonymization attacker probe under the \texttt{full} masked payload (the exact view the agent receives in \textit{blinded} mode). Each attacker sees 200 randomly drawn probes; rates are over successful responses only. Top-1 and joint success are the headline numbers (Section~\ref{sec:abl-deanon}); board accuracy is intentionally high (board field unmasked) and serves as a sanity check.}
\label{tab:deanon-full}
\end{table*}